\pdfoutput=1

\documentclass[11pt]{article}

\usepackage[final]{acl}

\usepackage{times}
\usepackage{latexsym}
\usepackage[most]{tcolorbox}
\usepackage{lipsum} 
\usepackage[T1]{fontenc}

\usepackage[utf8]{inputenc}

\usepackage{microtype}

\usepackage{inconsolata}

\usepackage{graphicx}

\usepackage{tcolorbox}
\usepackage{booktabs}
\usepackage{amsmath}
\usepackage{amsfonts}
\usepackage{tabularx}
\usepackage{multicol}
\usepackage{multirow}
\usepackage{makecell}
\usepackage{hyperref}
\usepackage{xurl}

\title{Breaking the Reviewer: Assessing the Vulnerability of Large Language Models in Automated Peer Review Under Textual Adversarial Attacks}

\author{
Tzu-Ling Lin\textsuperscript{1}, 
Wei-Chih Chen\textsuperscript{1}, 
Teng-Fang Hsiao\textsuperscript{1}, 
Hou-I Liu\textsuperscript{1}, 
Ya-Hsin Yeh\textsuperscript{1}, \\
\bf{Yu-Kai Chan}\textsuperscript{1}, 
\bf{Wen-Sheng Lien}\textsuperscript{1}, 
\bf{Po-Yen Kuo}\textsuperscript{1}, 
\bf{Philip S. Yu}\textsuperscript{2}, 
\bf{Hong-Han Shuai}\textsuperscript{1}\thanks{Corresponding author.} \\
\textsuperscript{1}National Yang Ming Chiao Tung University \\
\textsuperscript{2}University of Illinois Chicago 
}

\begin{document}
\maketitle
\begin{abstract}
Peer review is essential for maintaining academic quality, but the increasing volume of submissions places a significant burden on reviewers. Large language models (LLMs) offer potential assistance in this process, yet their susceptibility to textual adversarial attacks raises reliability concerns. This paper investigates the robustness of LLMs used as automated reviewers in the presence of such attacks. We focus on three key questions: (1) The effectiveness of LLMs in generating reviews compared to human reviewers. (2) The impact of adversarial attacks on the reliability of LLM-generated reviews. (3) Challenges and potential mitigation strategies for LLM-based review. Our evaluation reveals significant vulnerabilities, as text manipulations can distort LLM assessments. We offer a comprehensive evaluation of LLM performance in automated peer reviewing and analyze its robustness against adversarial attacks. Our findings emphasize the importance of addressing adversarial risks to ensure AI strengthens, rather than compromises, the integrity of scholarly communication. Our project page can be found in \url{https://github.com/Lin-TzuLing/Breaking-the-Reviewer.git}
\end{abstract}

\section{Introduction}
Peer review is crucial for examining the quality and integrity of academic publications. However, recent studies \cite{pan2018memory, fortunato2018science, bornmann2021growth, weissgerber2021automated, severin2021overburdening, shah2022challenge} highlighted a rapid increase in manuscript submissions, placing a significant burden on reviewers. Increasing specialization within fields further complicates the process, making it harder for reviewers to provide insightful comments on topics outside their expertise. Consequently, some reviews are less comprehensive or insightful than they might be with more time or specialized knowledge~\cite{shah2022challenge, shah2018design, severin2021overburdening, staudinger2024analysis}.

In response to these challenges, there is growing interest in leveraging automated systems to assist in the peer review process. Advances in natural language processing (NLP) and machine learning have opened promising avenues for developing tools that can augment human efforts in review generation. Early work by~\citet{nikiforovskaya2020automatic} proposed an automated review generation process that combines traditional bibliometric analysis with BERT-based extractive summarization to identify key contributions in manuscripts. More recently,~\citet{zhou2024llm, du2024llms} conducted a comprehensive evaluation of LLMs on automatic paper reviewing tasks, exploring the reliability of LLMs as potential reviewers.

Despite the benefits, the integration of LLMs into the peer review process raises concerns. One pressing issue is the vulnerability of LLM-generated reviews to textual adversarial attacks. Given that LLMs heavily rely on textual input, simple manipulations of the content could significantly distort their assessments. This vulnerability poses a security risk, especially if reviewers, intentionally or unintentionally, use mainstream LLMs to generate reviews without rigorous oversight.

In this paper, we investigate the robustness of LLMs as scientific paper reviewers. Specifically, we address three research questions:\\
\textit{(1) How effectively can LLMs generate reviews that approach the quality of human-written ones?}\\
\textit{(2) To what extent do textual adversarial attacks impact the reliability of LLM-generated reviews?}\\
\textit{(3) What challenges arise when applying textual adversarial attacks to LLM-based reviewing systems, and how can they be mitigated?}

To address these questions, we begin by evaluating the performance of leading LLMs, including closed-source models  (e.g., \texttt{GPT-4o}, \texttt{GPT-4o-mini}), as well as open-source models  (e.g., \texttt{Llama-3.3-70B}, \texttt{Mistral-small-3.1}), on two core peer review tasks: review generation and score prediction. Our evaluation demonstrates the models' capability to produce coherent and fluent reviews across multiple assessment aspects. We then apply common textual adversarial attack techniques to test the robustness of the LLM's evaluations. Our findings reveal significant vulnerabilities, as the LLM's assessments can be easily distorted through simple manipulations of the input text. Our contributions are threefold:
\begin{itemize}
    \item \textbf{Aspect-based evaluation of LLM reviewer:} We conduct an in-depth evaluation of LLM performance in aspect-based automatic paper reviewing, comparing outputs with human reviewers in the review generation task.
    \item \textbf{Adversarial robustness analysis:} We investigate the reliability of LLM-generated reviews against textual adversarial attacks in the score prediction task and propose a novel method, Attack Focus Localization, to pinpoint vulnerabilities in lengthy academic documents.
    \item \textbf{Limitations of existing defenses:} We analyze current defense methods and show that they struggle to detect adversarial manipulations or reliably identify LLM-generated reviews for securing the fairness of review systems.
\end{itemize}

\section{Related Work}

\subsection{Automated Paper Review-Related Tasks}
The exponential growth in academic publications has significantly increased peer reviewers' workload, straining traditional peer review systems \cite{weissgerber2021automated, severin2021overburdening, shah2022challenge}. To mitigate this burden, research in automated peer review has been developed and can be broadly divided into two streams: (i) pre-peer review screening tools and (ii) approaches leveraging advanced natural language processing (NLP) and large language models (LLMs).

Pre-peer review automation mainly focused on screening tasks that help ensure compliance with journal policies, such as readability assessment, plagiarism detection, and statistical error checking \cite{kilicoglu2018automatic, riedel2020oddpub, riedel2022replacing, zhang2010crosscheck, nuijten2016prevalence, checco2021ai, schulz2022future}. While effective in improving editorial efficiency, these tools are limited to surface-level checks and cannot evaluate a paper within its broader scientific context or assess methodological appropriateness.

The transition to NLP-based review generation reflects an effort to move beyond rule-based checks \cite{kuznetsov2024can}. For example, \citet{nikiforovskaya2020automatic} integrated co-citation analysis with BERT-based summarization to generate review content, while \citet{yuan2022can} developed a system using BART and established criteria for assessing review quality. Compared to earlier screening tools, these approaches move closer to approximating human-like judgment, but they remain constrained by domain specificity, dataset scale, and the reliability of generated evaluations.

Building on these early NLP-based pipelines, recent research has turned to large language models (LLMs). \citet{lu2024ai} introduced an LLM-driven scientific research and evaluation framework, \citet{du2024llms} investigated LLMs as (meta-)reviewers in NLP venues, and \citet{yu2024automated} proposed a framework combining review generation with self-critique. The emergence of GPT-4 further prompted several studies \cite{liu2023reviewergpt, robertson2023gpt4, zhou2024llm, liang2024can} that benchmarked LLM-generated reviews against human reviewers, highlighting both strengths (e.g., consistency, speed) and limitations (e.g., lack of in-depth domain expertise). In parallel, other works have examined the detectability of LLM-generated reviews \cite{wu2025survey, wu2024detectrl, yu2024your}, emphasizing the difficulty of avoiding false positives and negatives. Unlike earlier screening tools, these studies underscore the dual problem of harnessing LLMs for review assistance while safeguarding review integrity.

\subsection{Vulnerabilities of Large Language Models to Adversarial Text Attacks}

Adversarial text attacks subtly modify input content to mislead language models, resulting in altered outputs. Such attacks can be categorized along two dimensions: the granularity of perturbations and the attacker’s knowledge of the target model. At the perturbation level, character-level manipulations (insertion, deletion, swapping) can cause significant disruptions with minimal changes \cite{gao2018black, ebrahimi-etal-2018-hotflip, belinkov2017synthetic}, but generate nonsensical tokens and cause obvious typos. Word-level attacks, typically based on synonym replacement or contextual tweaks, preserve grammaticality yet can be easier to detect through lexical constraints  \cite{jin2020bert, li2020bert, maheshwary2021generating}. Sentence-level attacks leverage paraphrasing or syntactic restructuring, which can be highly deceptive though computationally intensive \cite{qi2021mind, qi2021hidden}. Attacks are also distinguished by attacker knowledge: white-box attacks exploit full model access, enabling gradient-based perturbations, whereas black-box attacks operate without internal information, trading precision for broader applicability in real-world scenarios.

While these techniques have long been studied in NLP, the rise of LLMs introduces new vulnerabilities. \citet{yao2024survey, kumar2024adversarial} highlight threats such as data poisoning, backdoor attacks, and prompt injection. Compared to traditional models, LLMs also exhibit behavioral weaknesses: position and verbosity biases, where models favor earlier options or longer responses \cite{liu2024lost, saito2023verbosity}, as well as self-enhancement bias, where models tend to portray themselves positively \cite{zheng2023judging}. These biases complicate evaluation, making LLMs vulnerable to adversarial inputs while simultaneously undermining their reliability as evaluators of their own outputs.

The impact of such vulnerabilities becomes particularly salient in high-stakes applications. In the context of peer review, \citet{robertson2023gpt4} observed that GPT-4 could detect inconsistencies after introducing adversarial modifications to abstracts, but it struggled with subtler manipulations such as shifts in formality. Similarly, \citet{raina2024llm} showed that adversarial manipulation could significantly inflate evaluation scores, raising concerns about the fairness and reliability of automated LLM-based reviewing processes.

\section{LLM Vulnerability on Paper Reviews}

\begin{figure*}[ht]
    \begin{center}
    \includegraphics[width=\textwidth]{Figures/EMNLP_framework_lowersolution.png}
    \caption{Illustration of a scenario where an LLM is under adversarial attack: the author may introduce specific patterns (e.g., typos) into the paper, causing the feedback generated by LLM reviewers to be misled.}
    \label{fig:framework}
    \end{center}
\end{figure*}

In response to the growing use of LLMs in peer review~\cite{zhou2024llm,tyser2024ai,yu2024your,lu2024ai}, we investigate their ability to automatically generate scientific reviews and examine their robustness against textual adversarial attacks, which are subtle edits that preserve meaning but may influence model outputs.

In Section~\ref{sec:peer_review_task_formulation}, we simulate a realistic review workflow in which reviewers employ LLMs in a zero-shot manner to produce aspect-tagged reviews and scores with minimal manual effort. To evaluate the resilience of this process, we introduce \emph{Attack Focus Localization} in Section~\ref{sec:lcs_match}, a method that identifies effective textual adversarial attack candidates in extremely long inputs via Longest Common Subsequence (LCS) matching. In Section~\ref{sec:adv_attack_methods}, we integrate this localization algorithm with existing adversarial perturbations (e.g., typos or pattern-based edits at the character, word, or sentence level), which can potentially bias the LLM’s output (see Fig.~\ref{fig:framework}).

\subsection{Peer Review Task Formulation}
\label{sec:peer_review_task_formulation}
To systematically evaluate the capabilities and robustness of LLMs in automated peer reviewing, we define two primary tasks that emulate essential components of the human review process: (1) Review Generation and (2) Score Prediction. These tasks are respectively designed to assess the LLM's ability to produce comprehensive written feedback and assign quantitative evaluations across predefined criteria. 

\noindent{\textbf{Task 1: Review Generation.}}
The most direct application of leveraging LLMs as reviewers is the automatic generation of reviews for academic papers. In this task, we investigate the capability of LLMs to produce constructive peer reviews that are coherent and consistent with human evaluations. Building on the aspect labels provided by \citet{yuan2022can}, we construct a predefined aspect typology aligned with the ACL review guidelines. This typology includes the categories: ``Summary'', ``Motivation'', ``Substance'', ``Originality'', ``Soundness'', ``Clarity'', ``Replicability'', ``Meaningful comparison'', and ``None''. Except for ``Summary'', each aspect is further classified into positive and negative sentiments to capture both strengths and weaknesses of the paper. The ``None'' tag is assigned to sentences that do not fit any categories.

\begin{figure}[t]   
    \begin{center}
    \includegraphics[width=0.48\textwidth]{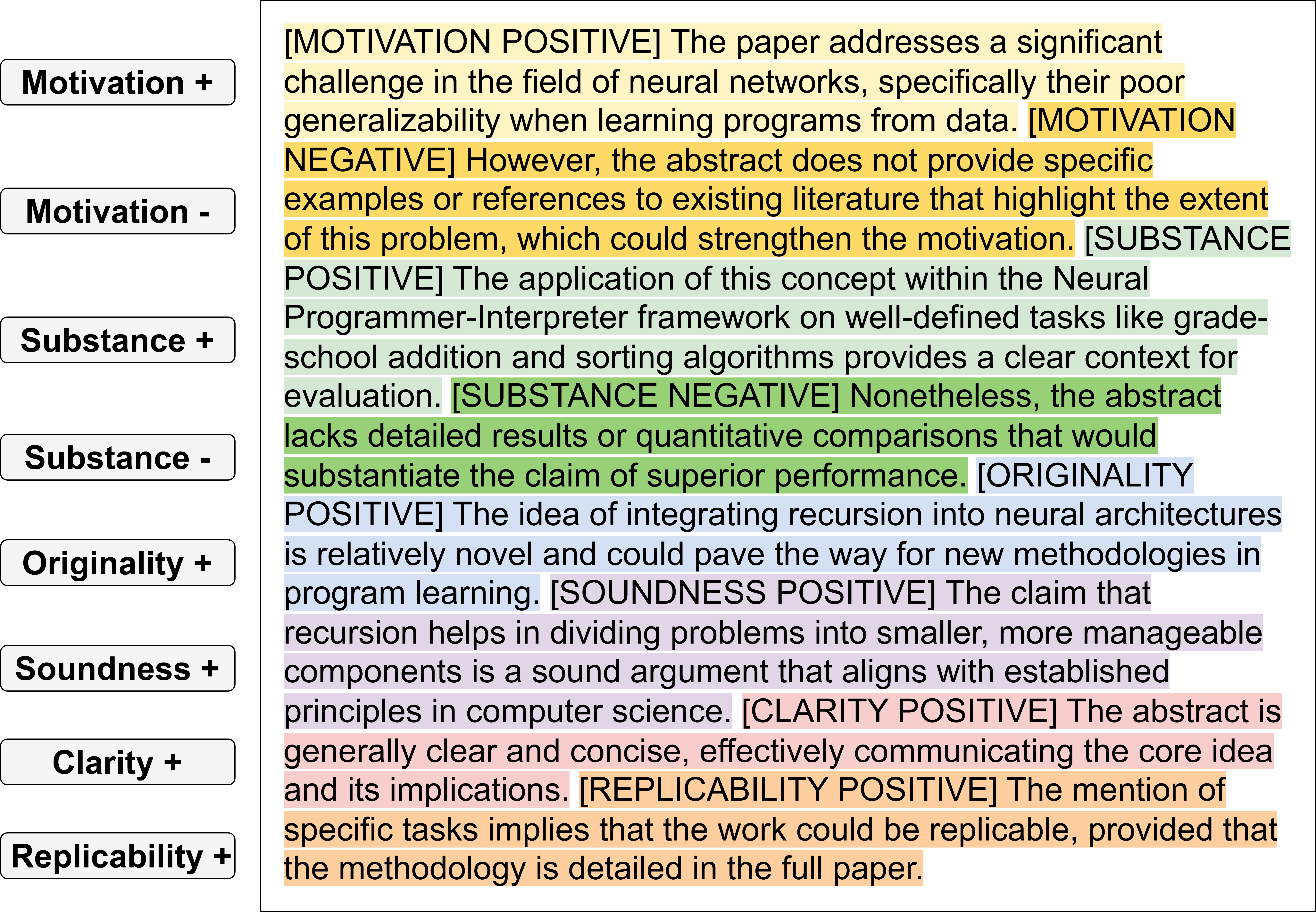}
    \caption{The example of aspect-tagged review generated by \texttt{GPT-4o-mini}.}
    \label{fig:example_review}
    \end{center}
\end{figure}

Using this typology, we prompt LLMs to write a review in the style of ICLR (International Conference on Learning Representations), instructing the model to tag each sentence with the corresponding aspect type at the beginning of the sequence. This approach ensures that the generated review addresses all relevant evaluation criteria and provides a structured assessment of the paper. An illustration of an automatically generated review with aspect tags is shown in Fig.~\ref{fig:example_review}.

\noindent{\textbf{Task 2: Score Prediction.}}
Score prediction serves as a vital complement to review generation by offering quantitative evaluations of a paper across multiple aspects~\cite{kang-etal-2018-dataset, zhou2024llm}. Its structured and numerical format provides clearer evaluation standards and enables more consistent comparisons. In this task, we assess the capability of LLMs to assign scores from different perspectives, simulating the expert judgments typically made by human reviewers.

Following the setup from~\citet{zhou2024llm}, we prompt the LLM to predict scores for predefined aspects: ``overall,'' ``substance,'' ``appropriateness,'' ``meaningful comparison,'' ``soundness correctness,'' ``originality,'' ``clarity,'' and ``impact.'' Each score is expected to fall within a specific range (e.g., 1 to 10), in accordance with the scoring conventions of the target conference. This setup allows us to evaluate not only the LLM’s understanding of the paper but also its ability to apply evaluation criteria in a consistent and contextually appropriate manner.

To provide additional insight into the model’s reasoning, we ask it to include a brief justification for each score (as illustrated in Appendix~\ref{sec:case_study}). This practice mirrors that of human reviewers, who often accompany their ratings with explanatory comments. By comparing the LLM-generated scores and justifications with those provided by human reviewers, we can assess the alignment between automated and expert evaluations. Furthermore, this task enables us to study the impact of adversarial text perturbations on the model’s scoring behavior. By introducing subtle modifications to the input, we can observe how the LLM’s assigned scores shift across different aspects. This analysis reveals the model’s sensitivity to input variation and its robustness in maintaining consistent evaluations under potentially manipulated conditions.

\subsection{Attack Focus Localization}
\label{sec:lcs_match}
In paper reviewing tasks, the content under review is often too lengthy, making it computationally infeasible to directly search for adversarial patterns across the entire paper. To address this challenge, we first identify the content areas of the paper that attract the most attention from the LLM while generating reviews by feeding the clean paper content into the model and requesting a review.

Following this, we apply the Longest Common Subsequence (LCS) algorithm to identify matching subsequences between the paper content and the generated review. These matched subsequences are then highlighted as potential modifiable content for adversarial pattern injection.
Let $x_{clean}$ represent the clean content of a single paper. For each paper, we extract a set of modifiable content by finding matching subsequences using LCS between the paper's clean content and the review generated by the LLM. The process can be formalized as follows:

\noindent{\textbf{1. Identifying Modifiable Indexes.}} The modifiable subsequences are indexed by their start and end positions in the clean content:
\begin{equation*} \label{eq1}
\begin{aligned}
\mathcal{M}
& = 
\text{LCS}\left(x_{\text{clean}}, f_r(x_{clean}) \right)\\
& = 
\{m_1, m_2,\dots, m_N \},
\end{aligned}
\end{equation*}

\noindent where $\mathcal{M}$ is the modifiable sequences set for a single paper, $f_r(\cdot)$ denotes the review generation by LLM $f$ based on the input content. \text{LCS} identifies the longest common sequence matching algorithm. \footnote{LCS can be character-level or word-level depending on the attack method; implementation details could be found in Appendix \ref{app:LCS_split}.}, $m$ represents modifiable subsequences, and $N$ is the total number of matched subsequences. 

\noindent{\textbf{2. Searching Optimal Adversarial Pattern.}}  The objective of these attacks is to find perturbations that lead to the maximum score shift for each paper, mathematically formulated as:
\[
\mathcal{P}_m = A \left( x_{clean}, m \right)\ , \forall m \in \mathcal{M}
\]
\begin{equation*}
\resizebox{0.5\textwidth}{!}{
$
X_{\text{adv}} = \left\{ \left( x_{\text{clean}} \setminus m \right) \cup \left( m + \delta \right) \mid \forall m \in M, \forall \delta \in P_m \right\}
$
}
\end{equation*}
\[
x_{adv}^* = \arg \max_{x \in \mathcal{X}_{adv}} \left(f_s(x_{adv} - f_s(x_{clean})) \right)
\] where $\mathcal{P}_m$ defines the possible set of perturbations for every modifiable sequence $m$, identified through the transformations specified by the attack method $A$. $\delta$ represents the textual perturbation applied to the original content, and $x^{*}_{adv}$ is the optimal perturbed content found through the search method of the attack. $f_s(\cdot)$ represents the sum of the score for all aspects predicted by LLM $f$.

The ultimate goal of the attack is to maximize the total score shift across all papers, expressed as:
\[
\text{Total Score Shift} = \sum_{x \in \mathcal{D}}^{} \left( f_s(x^{*}_{adv}) - f_s(x_{clean}) \right)
\]
where $\mathcal{D}$ represents the dataset of papers.

\subsection{Textual Adversarial Attacks}
\label{sec:adv_attack_methods}
To examine the robustness of both review generation and score prediction tasks, we investigate a set of representative textual adversarial attack methods. We categorize these methods according to common strategies in textual adversarial attacks. For each strategy, we introduce widely recognized approaches in black-box settings to evaluate the robustness limits of LLMs.

\noindent{\textbf{Character-Level Attacks.}}
For character-level adversarial attacks, we employ \textcolor{gray}{\texttt{DeepWordBug}}~\cite{gao2018black} and \textcolor{gray}{\texttt{PuncAttack}}~\cite{formento2023using}. These methods mainly involve introducing typographical or grammatical errors in words by adding, deleting, repeating, replacing, or swapping characters or punctuation marks. Such perturbations can significantly affect the model's performance while being less noticeable to humans.

\noindent{\textbf{Word-Level Attacks.}}
For word-level adversarial attacks, we implement \textcolor{gray}{\texttt{TextFooler}}~\cite{jin2020bert} and \textcolor{gray}{\texttt{BERT-Attack}}~\cite{li2020bert} to deceive the LLM. Both attacks focus on replacing words with their synonyms or similar terms in a contextualized embedding space, maintaining semantic similarity while altering the input text. 

\noindent{\textbf{Sentence-Level Attacks.}}
For the sentence-level attacks, we use \textcolor{gray}{\texttt{StyleAdv}}~\cite{qi2021mind} as representatives for sentence-level attacks. These methods involve transforming sentences in the paper content into specific text styles and modifying stylistic elements without altering the semantic meaning. Such transformations are designed to be imperceptible to human readers but can potentially mislead the LLM.
    

To conclude, applying these adversarial attack methods enables us to assess the robustness of LLMs in automated peer review. By introducing perturbations at the character, word, and sentence levels, we identify vulnerabilities that compromise the LLM's reliability. The degree of adversarial changes and examples from different attacks are provided in Appendix~\ref{app:attackExample_modificationRate}, which offers a visualization of how these perturbations alter the original sentences. Also, the algorithm of textual adversarial attacks can be found in Appendix~\ref{app:attack_implementation}.

\section{Experiment}
\subsection{Experiment Settings}
\noindent\textbf{LLM Selection and Configuration.}
We employ both closed- and open-source LLMs for automated peer review. From the closed-source family, we use \texttt{GPT-4o} and its cost-efficient variant \texttt{GPT-4o-mini}, both of which have shown strong performance in text evaluation tasks \cite{robertson2023gpt4, zhou2024llm}. For open-source models, we adopt \texttt{Llama-3.3-70B} for its general language capabilities and \texttt{Mistral-small-3.1} for its competitive performance with \texttt{GPT-4o-mini}, particularly in long-context understanding. Detailed model configurations and prompt templates are provided in Appendix~\ref{app:parameters} and~\ref{app:example_prompt}.

\noindent\textbf{Dataset.}
PeerRead~\cite{kang-etal-2018-dataset} is one of the earliest publicly released datasets of peer reviews in the NLP community. Following~\citet{zhou2024llm}, we use the ICLR-2017 subset for both review generation and score prediction tasks. This subset contains 427 papers (177 accepted and 255 rejected), with reviews written by experts and aspect-specific scores assigned by annotators. Experimental results on adversarial attack performance and score shift comparison are provided in Appendix~\ref{app:peerread_attack_result}.

AgentReview~\cite{jin-etal-2024-agentreview} presents a large-scale dataset curated through their LLM-based peer review simulation framework. It covers ICLR papers from 2020–2023, sourced from OpenReview, and includes 523 papers (350 rejected, 125 poster, 29 spotlight, and 19 oral).

Since the reviews in both datasets are not manually annotated with aspect tags, we leverage the aspect tagger pre-trained on the ASAP dataset \cite{yuan2022can} to automatically provide aspect tags for human-written reviews.

\subsection{Evaluation Metrics}
\noindent\textbf{Review Quality Metrics.} 
To address the research question of \textit{how effective LLMs are in generating coherent and insightful reviews compared to human reviewers}, we conduct a comparison between LLM-generated reviews and human-annotated reviews, with the latter serving as ground truth. Following the metrics proposed in prior work~\cite{yuan2022can, du2024llms, yu2024automated}, we evaluate the quality of automated reviews across multiple dimensions. Our evaluation assesses both the overall quality of LLM-generated reviews and their consistency with human judgments:

\noindent (1) \textit{Comprehensiveness}: 
For automatically generated reviews, we expect them to cover diverse aspects of the paper. We calculate the Aspect Coverage (ACOV) metric to assess the generated review. In particular, the number of aspects in a predetermined aspect typology that a review has covered can be used to quantify aspect coverage.
\[
ACOV = \frac{\text{\# of aspect tags in generated review}}{\text{\# of predefined aspects}}
\]

\noindent (2) \textit{Similarity to human}: 
To evaluate the semantic equivalence between LLM-generated reviews and multiple human-annotated ground truth reviews, we utilize classical statistical methods: ROUGE-1, ROUGE-2, ROUGE-L \cite{lin2004rouge}, and the BertScore metric \cite{zhang2019bertscore}. Rather than averaging scores across all human references, we select the maximum score between the LLM-generated review and the human-written ground truths. This approach assumes that closely aligning with every reference is unnecessary; instead, identifying the best match offers a more meaningful assessment of semantic equivalence. Although manually annotated reviews are not free from bias and omissions, human-written evaluations can still be regarded as a practical gold standard.

While these automatic metrics capture surface-level qualities such as fluency and lexical similarity to human-written text, they offer limited insight into deeper dimensions, including coherence, usefulness, and critical insight. To address this gap, we conduct a controlled human evaluation to complement the automatic analysis, as described in Appendix~\ref{app:human_evaluation_review_quality}.

\noindent{\textbf{Robustness Metrics.}} To quantitatively assess the robustness of the LLM in the face of adversarial attacks, we define the following metrics.

\noindent (1) \textit{Sensitivity Analysis}: 
    Building on prior work in adversarial attacks, we adopt two straightforward metrics to evaluate the robustness of LLMs as reviewers: Attack Success Rate (ASR) and Score Shift. An attack is considered successful if the score shift between the clean output and the attacked output is at least +1.0 in the total score. ASR is defined as the proportion of such successful attacks, providing a clear indicator of the LLM's vulnerability under adversarial conditions.

\noindent (2) \textit{Statistical Measures}: 
    We use the Wilcoxon signed-rank test \cite{woolson2005wilcoxon} as the statistical support. With the p-value set to 0.05, we pair the score that is predicted based on clean content and attacked content and test if the hypothesis that there exists a significant positive score shift after applying an adversarial attack holds true.

In addition to the quantitative results, we conducted a user study to assess the stealthiness of attacks in a realistic peer review setting. Specifically, we evaluated participants’ ability to identify attacked papers and localize the perturbed paragraphs. The detailed settings and full user study results are reported in Appendix~\ref{app:human_evaluation_attack_imperceptibility}.

\subsection{Evaluation -- Review Quality}
\label{sec:evaluation_review_quality}
Table \ref{tab:quality_result_agentreview} reports automatic evaluation results for the quality of reviews generated by LLMs.
To establish comparison baselines, we first categorize human-written reviews into two types: \textbf{(1) Within-Paper Pairs}: review pairs originating from the same paper, capturing the semantic similarity of reviews written for the same context. These pairs are generated using all possible combinations of reviews for each paper (e.g., $C^n_2$ combinations for $n$ reviews per paper). \textbf{(2) Across-Paper Pairs}: review pairs formed from different papers, reflecting the semantic similarity of reviews written for unrelated content. This set is constructed by enumerating all possible combinations of reviews across the entire dataset ($C^n_2$ combinations for N total human-written reviews) and excluding those counted as within-paper pairs.\footnote{Since Across-Paper Pairs are substantially more numerous, we randomly sample 10\% for similarity calculation to reduce computational cost. Unlike similarity-based metrics, ACOV does not rely on paired data, we do not split it into within- and across-paper categories. Consequently, its score for human-written reviews is identical across pair types.} We observe that:\\
\noindent\textbf{The ACOV of LLM-generated reviews is higher than human-written ground truths.} The significance is further proved by the Wilcoxon signed-rank test. The result can be attributed to the structured review guidelines provided in the prompt, which guide the model in covering a more comprehensive range of aspects. 
These findings suggest that, under explicit guidance, LLMs can provide broader perspectives on review principles, covering both positive and negative sentiments while focusing on the desired qualities of reviews. 
Importantly, our intention is not to claim that LLM-generated reviews surpass human reviews, but rather to highlight their potential for offering evaluations across multiple perspectives. 
We also note that the ACOV decreases when the scoring aspects are not explicitly specified, due to the lack of direct alignment. 
A detailed interpretation is provided in Appendix~\ref{app:abs_acov}.\\
\noindent\textbf{Within-paper pairs exhibit higher ROUGE scores.} Reviews for the same paper consistently achieve higher R-1, R-2, and R-L scores, reflecting shared vocabulary, content, and structural similarity due to discussing the same paper-specific details. In contrast, across-paper pairs naturally show lower scores given their divergent content. LLM-generated reviews also show this trend: while slightly lower than human within-paper pairs, their ROUGE scores remain substantially higher than across-paper pairs, indicating the model’s ability to integrate paper-specific information and generate coherent and contextually aligned reviews.\\
\noindent\textbf{Across-paper pairs exhibit BERTScore values comparable to within-paper pairs.} This may be attributed to the semantic embeddings capturing shared linguistic structures or standardized grading language commonly used by reviewers, despite differences in paper content. For LLM-generated reviews, the similarity in BERTScore across both pair types indicates that LLMs are capable of generating reviews with structural and linguistic characteristics similar to those of human reviewers.

\begin{table}[ht]
\fontsize{7.1}{11}\selectfont
\caption{Evaluation of LLM-generated reviews' Comprehensiveness and Similarity to humans of AgentReview~\cite{jin-etal-2024-agentreview} dataset. ACOV denotes Aspect Coverage. R-1, R-2, R-L, and BS indicate the Rouge-1/2/L, and BertScore, respectively. \textbf{Human (within)} reports the mean scores of within-paper pairs, \textbf{Human (across)} reports the mean over across-paper pairs.}
\label{tab:quality_result_agentreview}
\begin{tabularx}{0.5\textwidth}{Xccccc}
\toprule
Reviewer & ACOV & R-1 & R-2 & R-L & BS \\
\cmidrule(lr){1-1}\cmidrule(r){2-6}
Human (within) & \textbf{0.4980} & \textbf{0.3746} & \textbf{0.0774} & \textbf{0.1813} & \textbf{0.8440} \\
Human (across) & \textbf{0.4980} &  0.1994 & 0.0292 & 0.1069 & 0.8163 \\
\cmidrule(lr){1-1}\cmidrule(r){2-6}
GPT-4o-mini & 0.6111 & 0.3185 & 0.0670 & 0.1594 & \textbf{0.8285} \\
GPT-4o & \textbf{0.9162} & 0.3134 & 0.0666 & 0.1574 & 0.8254 \\
Llama-3.3-70B & 0.6594 & 0.3127 & \textbf{0.0750} & \textbf{0.1666} & 0.8272 \\
Mistral-small-3.1 & 0.9149 & \textbf{0.3204} & 0.0731 & 0.1552 & 0.8226 \\
\bottomrule
\end{tabularx}
\end{table}

\subsection{Evaluation -- Attack Effectiveness}
\begin{table*}[tb]
\caption{Adversarial Attack Performance and Score Shift Comparison of AgentReview~\cite{jin-etal-2024-agentreview} dataset. ``ASR'' represents the Attack Success Rate. ``Avg. Score shift'' signifies the average score shift between attacked and clean outputs. ``Avg. \#Pos/Neg Shift'' indicates the average change in the number of positive and negative tags assigned to the reviews. ``\#Queries'' refers to the average number of times LLM is queried. The values are reported as result from \texttt{GPT-4o-mini} / \texttt{GPT-4o} / \texttt{Llama-3.3-70B} / \texttt{Mistral-small-3.1}.}
\label{tab:adv_result}
\small
\begin{center}\
\resizebox{\textwidth}{!}{%
\setlength{\tabcolsep}{5pt}
\begin{tabular}{cccccc}
\toprule
\multirow{2}{*}[-3pt]{Attacks} &
\multirow{2}{*}[-3pt]{ASR $\uparrow$} &
\multicolumn{3}{c}{Average Shift} &
\multirow{2}{*}[-3pt]{\#Queries $\downarrow$} \\
\cmidrule(lr){3-5}
 &  &  
 Score $\uparrow$ & 
 \#Pos Tags $\uparrow$ & 
 \#Neg Tags $\downarrow$ & 
 \\
\cmidrule(lr){1-1}\cmidrule(r){2-6}
DeepWordBug 
& 0.87 / 0.90 / 0.81 / 0.91  
& 3.44 / 4.41 / 2.56 / 3.55 
& 0.24 / \textbf{0.01} / -0.08 / -0.02
& -0.78 / -1.49 / -1.70 / -0.32 
& 43 / 51 / 52 / 55  \\
PuncAttack 
& 0.85 / 0.87 / 0.77 / 0.87  
& 3.44 / 4.05 / 2.52 / 3.59  
& 0.27 / \textbf{0.01} / -0.11 / \textbf{-0.01} 
& -0.82 / -1.12 / -1.79 / -0.19 
& 51 / 64 / 66 / 69 \\
TextFooler 
& \textbf{0.89} / \textbf{0.92} / \textbf{0.85} / 0.91 
& \textbf{3.77} / 4.71 / \textbf{2.98} / 3.94 
& 0.33 / 0.00 / -0.15 / -0.03  
& \textbf{-0.84} / -1.23 / -2.11 / -0.37  
& 47 / 52 / 58 / 58 \\
BERT-Attack 
& 0.87 / 0.89 / 0.71 / 0.90  
& 3.59 / 4.65 / 2.36 / \textbf{4.17}  
& \textbf{0.36} / -0.01 / \textbf{-0.06} / -0.03
& -0.69 / \textbf{-1.52} / -1.51 / \textbf{-0.47}
& 54 / 56 / 56 / 66  \\
StyleAdv 
& \textbf{0.89} / \textbf{0.92} / 0.82 / \textbf{0.92} 
& 3.62 / \textbf{4.73} / 2.87 / 3.97 
& 0.30 / \textbf{0.01} / -0.15 / -0.04 
& -0.82 / -1.31 / \textbf{-2.25} / -0.38
& \textbf{25} / \textbf{32} / \textbf{25} / \textbf{30} \\
\bottomrule
\end{tabular}%
}
\end{center}
\end{table*}

To assess the robustness of LLM-based reviewers, we employ widely recognized textual adversarial attacks (described in Section~\ref{sec:adv_attack_methods}) to manipulate the content of input paper. Table~\ref{tab:adv_result} summarizes the results of these attack baselines aimed at artificially inflating the review scores predicted by LLMs. We observe that:\\
\noindent\textbf{Most attacks achieve a high attack success rate (ASR).} ASR among different granularities of attacks ranging from 71\% to 92\%, accompanied by substantial score shifts. These findings demonstrate the effectiveness of adversarial methods in deceiving LLM-based reviewers and highlight the models' sensitivity to such subtle manipulations injected into a lengthy document.\\
\noindent\textbf{Changes in the distribution of review sentiment tags.} Specifically, the average increase in positive tags (\textit{Avg. \#Pos Shift}) correlates with higher positive sentiment scores after attacks, while the average number of negative tags (\textit{Avg. \#Neg Shift}) consistently decreases. This suggests that even minor content perturbations can bias LLMs toward generating more favorable feedback, suppressing critical evaluations. As the predicted scores inflate, the models tend to produce more positive assessments while reducing the expression of negative ones, leading to overly optimistic reviews that lack the balanced critiques typical of human reviewers.\\
\noindent\textbf{Negative tag suppression aligns with inflated scores.} Although positive tags sometimes decrease after attacks, the reduction in negative tags is generally more significant. This trend is particularly evident in the \texttt{Llama-3.3-70B} and \texttt{Mistral-small-3.1}, where adversarial modifications substantially reduce the number of negative aspect tags. One possible explanation is that these models inherently favor positive or neutral sentiment generation, making them more susceptible to adversarial manipulations that suppress negative feedback. This imbalance suggests that adversarial attacks effectively weaken the models' capacity to generate critical-depth reviews that are supposed to be provided in the peer-reviewing process.

To statistically validate the impact of adversarial attacks, we conducted a Wilcoxon signed-rank test to assess whether post-attack scores significantly exceed pre-attack scores. As shown in Table~\ref{tab:adv_result}, the average score shifts are statistically significant, with p-values consistently below 0.05. These results confirm that the observed score inflation is not due to random variation but is a direct consequence of the applied adversarial manipulations.

\begin{figure}[t]   
    \begin{center}
    \includegraphics[width=0.48\textwidth]{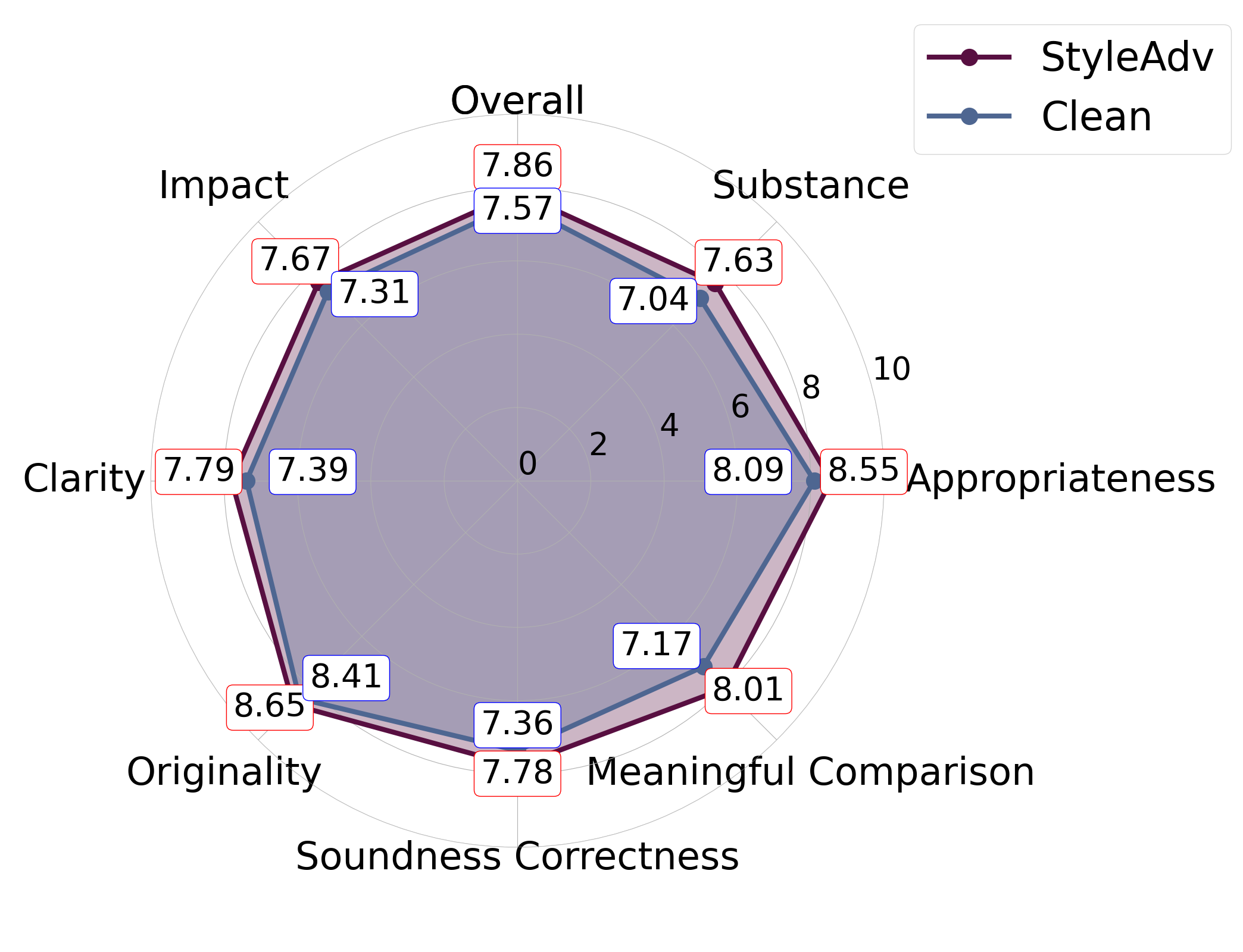}
    \caption{visualization of the average score of each aspect predicted by \texttt{GPT-4o-mini}. Clean denotes the prediction when input is not manipulated by \textcolor{gray}{\texttt{StyleAdv}}.}
    \label{fig:radar_chart}
    \end{center}
\end{figure}

Fig. \ref{fig:radar_chart} visualizes the average scores across various aspects in the predefined typology, comparing the results before and after applying \textcolor{gray}{\texttt{StyleAdv}} to the input paper content. The radar chart shows that post-attack scores generally surpass pre-attack scores across most aspects, highlighting the attack's effectiveness. Notably, some aspects, such as originality, clarity, and appropriateness, appear less sensitive to stylistic changes, likely because modifying writing style alone does not significantly alter the semantic meaning of the content. On the other hand, aspects like substance and meaningful comparison show greater vulnerability, as stylistic manipulation may reduce the depth of supporting evidence or coherence in comparisons, thus impacting the prediction more easily.

\section{Defenses and Limitations}
\label{sec:detection_limitations}
While recent studies have proposed using AI-generated text detectors (e.g., GPTZero, OpenAI Classifier) to identify LLM-authored reviews or adversarially manipulated inputs, such detection-based defenses face fundamental limitations when confronting the vulnerabilities in the peer-review process revealed in this work. 
\subsection{Detecting Adversarial Manipulations.} To apply detectors like GPTZero directly to the paper content to identify whether it has been adversarially manipulated (e.g., injected with typos or stylistic changes to mislead LLM reviewers) is beyond the current capabilities of AI-text detectors. These detectors are typically trained to distinguish between human-written and LLM-generated text, rather than to identify subtle, semantics-preserving perturbations designed to mislead LLM reviewers. These perturbations are designed to evade both human perception and statistical detectors, as confirmed by our human evaluation results. Thus, applying existing detectors to the paper content is unlikely to surface such attacks, and developing dedicated perturbation-aware detectors remains an important future direction.
\subsection{Detecting Machine-Generated Reviews.} A plausible use of detectors is to identify LLM-generated reviews, with the goal of adjusting confidence scores or triggering human audits. However, this strategy suffers from several challenges. \\
\noindent\textbf{Policy Ambiguity.}
Due to current conference policies (e.g., the ACL Policy on Publication Ethics\footnote{\url{https://www.aclweb.org/adminwiki/index.php/ACL_Policy_on_Publication_Ethics\#Guidelines_for_Generative_Assistance_in_Authorship}} and the NeurIPS 2025 LLM Policy\footnote{\url{https://neurips.cc/Conferences/2025/LLM}}) explicitly allowing the use of LLMs for language refinement, there is no clear boundary between LLM-written and LLM-assisted reviews. Applying a strict detection threshold risks unfairly penalizing legitimate human-written reviews that simply use LLMs to improve wording and fluency. \\
\noindent\textbf{Ease of Bypassing Detection.}
Existing detectors are vulnerable under various real-world application scenarios. An adversary can paraphrase LLM-generated reviews~\cite{krishna2023paraphrasing} or design prompts to guide LLMs in generating human-like text to bypass detectors~\cite{wu2024detectrl}, while these reviews still rely on LLM content that lacks a deep understanding of the paper. This undermines the goal of ensuring high-quality reviews.
Furthermore, as conferences begin to officially adopt LLM tools in the review pipeline (e.g., AAAI’s AI-powered peer review assessment system\footnote{\url{https://aaai.org/aaai-launches-ai-powered-peer-review-assessment-system/}}), relying solely on detector-based filtering appears inadequate to address the vulnerabilities introduced by adversarial manipulations of the paper content.

These challenges underscore the vulnerability of LLM-assisted review systems. The adversarial attacks are not only effective but also difficult to defend against using current methods.

\section{Conclusion}
We present a framework to evaluate the capabilities and robustness of large language models (LLMs) in automated peer-review tasks, addressing the growing burden on reviewers. Using mainstream LLMs, we perform review generation and score prediction, demonstrating their ability to provide coherent, insightful feedback. To assess robustness, we apply adversarial techniques to manipulate paper content and observe their impact on predicted scores, revealing vulnerabilities where subtle input changes affect reliability. We also propose \emph{Attack Focus Localization} to identify document regions most vulnerable to attacks, offering finer-grained insights into LLM weaknesses. Although such manipulations remain difficult to detect, exploring mitigation strategies—such as inference-time monitoring or human-in-the-loop verification—represents a promising avenue to improve the trustworthiness of LLM-assisted peer review.

\section*{Acknowledgments}
This work was supported in part by the National Science and Technology Council (NSTC) of Taiwan under Grant NSTC-112-2221-E-A49-094-MY3, and in part by the U.S. National Science Foundation (NSF) under Grants III-2106758 and POSE-2346158.

\section{Limitation}
Despite the valuable insights provided by this study, several limitations should be acknowledged. First, the evaluation of the LLM's performance was limited to the chosen representative models (i.e., \texttt{GPT-4o}, \texttt{GPT-4o-mini}, \texttt{Llama-3.3-70B}, and \texttt{Mistral-small-3.1}), and the generalizability of the findings to other LLMs remains uncertain. Different models may exhibit varying levels of robustness and accuracy, and future work should explore a broader range of models. Additionally, the scope of textual adversarial attacks applied in this research was restricted to certain common techniques. More sophisticated or domain-specific adversarial methods (e.g., adaptive attacks, non-black-box attacks) could further challenge the LLM's performance, which was not considered in this study. Finally, the focus on review generation and score prediction leaves out other critical aspects of the peer review process, such as the interaction between reviewers and authors or the integration of LLM-generated reviews into existing peer review workflows, which warrant further investigation.

\section{Ethical Considerations}
\label{sec:ethical_consideration}
This work does not rationalize or encourage the use of LLMs in the paper reviewing process. Instead, our aim is to provide a comprehensive discussion on the generation patterns of LLM-based reviewers and raise awareness of the vulnerabilities in applying LLM-generated results, as they can be easily manipulated through simple textual attacks and may lack the expertise and proper judgment required for such tasks.

\noindent\textbf{Bias and Fairness.} \quad LLMs are prone to biases based on their training data and also potentially suffer from position, verbosity, and self-enhancement biases, which may affect their fairness in paper review tasks. Ethical guidelines suggest conducting rigorous testing and mitigation strategies to reduce bias, ensuring that the reviews or scores generated are not distorted by manipulations in textual content that are introduced intentionally or not. Human reviewers should remain responsible for the final review and decision, even if LLMs assist in the process. Ethical AI usage involves maintaining human oversight to ensure accountability and avoid transferring full decision-making power to an algorithm that may lack context or moral judgment.
\textbf{AI Usage Compliance.} \quad Our research testing the robustness of LLM reviewers to adversarial attacks should adhere to ethical standards of security and integrity. This involves ensuring that such testing is conducted in a controlled environment, with clear goals of improving AI reliability rather than exploiting vulnerabilities for harmful purposes. Such experiments aim to strengthen the resilience and trustworthiness of AI systems.

\noindent\textbf{Fragility of AI-text Detectors.} \quad 
While utilizing AI-text detectors to identify machine-generated content appears to be a potential defense against our proposed threats, such mechanisms are insufficient to ensure security in real-world peer reviewing scenarios for several reasons (detailed in Section~\ref{sec:detection_limitations}). 

First, current detectors are primarily trained to distinguish between human-written and LLM-generated text, but they lack the capability to detect subtle, semantics-preserving adversarial perturbations in paper content.
Second, zero-shot detection of LLM-generated reviews can be trivially bypassed through paraphrasing or minor editing, allowing AI-text to evade filtering. Moreover, the increasing integration of LLMs into legitimate writing and reviewing workflows—often under ambiguous policy guidelines—renders binary detections ethically problematic and operationally difficult.

From an ethical standpoint, it is essential to critically assess that blanket reliance on such tools may foster false confidence while unintentionally marginalizing non-native English speakers who use LLMs for language assistance.
While AI-text detectors may serve as a useful complementary measure, they are not by themselves a sufficient defense against the challenges we uncover.

In all cases, human oversight remains essential for ensuring the fairness, accountability, and integrity of peer review.

\clearpage
\bibliography{custom}
\clearpage
\appendix

\section{Case Study}
\label{sec:case_study}
Table~\ref{tab:example_explanation} presents aspect-wise scores predicted by \texttt{GPT-4o-mini} along with corresponding explanations. With access to the full paper content, the model generates fluent and coherent justifications that highlight both strengths (Substance, Soundness, Originality) and areas for improvement (Clarity, Appropriateness). These explanations enhance the interpretability of the scoring process and demonstrate the model’s ability to reason about multiple dimensions of scholarly quality. Despite full access to the paper, the predicted scores still cluster within a relatively narrow range (6–8), indicating a limited critical depth. This suggests that LLM may default to generally favorable assessments unless explicitly prompted to be more discriminative.
Overall, the generated explanations not only offer transparency into the model’s evaluative reasoning but also reveal opportunities for further alignment with human reviewing norms.

\begin{table}[hbt]
\scriptsize
\caption{Predicted aspect scores from \texttt{GPT-4o-mini} and their corresponding explanations.}
\setlength{\tabcolsep}{3pt} 
\begin{center}
\renewcommand{\arraystretch}{1.5} 
\begin{tabular}{ccm{4.8cm}}
\toprule
\textbf{Aspect} & \textbf{Score} & \textbf{Explanation} \\
\cmidrule(lr){1-3}
Overall & 7.0 &
The paper makes significant contributions to GAN theory but has areas for improvement in clarity and practical implications. \\
Substance & 8.0 &
The rigorous theoretical analysis and proofs are strong, providing a solid foundation for the claims. \\
Appropriateness & 6.0 &
The motivation is relevant, but the practical implications could be better articulated for broader impact. \\
\makecell{Meaningful \\ Comparison} & 7.0 &
The paper contextualizes its contributions well but could engage more critically with alternative approaches. \\
\makecell{Soundness \\ Correctness}& 8.0 & 
The theoretical framework is robust, though some assumptions may limit applicability. \\
Originality & 7.0 &
The introduction of new tools is commendable, but the paper could better distinguish its contributions from existing work. \\
Clarity & 6.0 &
While generally clear, some sections are dense and could be simplified for accessibility. \\
Impact & 7.0 &
The findings have the potential to influence future research directions in GAN training, though practical applications need clearer emphasis. \\
\bottomrule
\end{tabular}
\end{center}
\label{tab:example_explanation}
\end{table}

\section{Granularity-Specific LCS Splitting}
\label{app:LCS_split}
The granularity of modifiable content found by the LCS algorithm is dynamic, as we split both the clean paper content and the generated review into segments based on the chosen attack level, which can be character-level, word-level, or sentence-level. The LCS matching is then performed at the appropriate granularity.

\noindent\textbf{Character-level Splitting.} The LCS algorithm operates at a fine-grained character level, enabling precise modifications that are particularly useful for attacks requiring small, detailed changes.

\noindent\textbf{Word-level Splitting.} We split both the paper content and the generated review into word units and constrain the LCS matching to operate on those units. Since word-level attacks typically involve synonym substitutions or similar changes in word granularity, this ensures modifiable subsequences correspond to complete words and that replacements do not fragment tokens.

\noindent\textbf{Sentence-level Splitting.} For sentence-level attacks, we first apply LCS at the word level to find matched subsequences between the clean content and the review, then extend each match bidirectionally to capture the full sentence containing the sequence. This ensures adversarial patterns are injected into complete sentences while preserving the overall syntactic and semantic coherence.

\section{Magnitude of Changes Across Different Adversarial Attacks}
\label{app:attackExample_modificationRate}
To quantify the magnitude of changes when applying different textual adversarial attacks on paper content, we report two standard metrics: \textbf{Modification Rate} and \textbf{Semantic Similarity}. 

Given a pair of paper content—original \( x_{\text{clean}} \) and adversarial \( x_{\text{adv}} \)—we first extract the subset of sentences that differ using sentence-level alignment and comparison. This results in a filtered set of sentence pairs:

\begin{align*}
\mathcal{S} = \big\{(s_{\text{clean}}, s_{\text{adv}})\ \big|\ 
& s_{\text{clean}} \in x_{\text{clean}},  s_{\text{adv}} \in x_{\text{adv}},
\nonumber\\
& \ s_{\text{clean}} \ne s_{\text{adv}} \big\}
\end{align*}

Each pair in $\mathcal{S}$ consists of an original sentence and its corresponding adversarial variant. The following two metrics are computed over this set:\\
\noindent\textbf{Modification Rate.} This metric measures the extent of perturbation introduced into the original text.
For each sentence pair in $\mathcal{S}$, the normalized character-level change is:

\begin{equation*}
\text{Modification Rate} = \frac{\Delta_\text{char}(s_{\text{clean}}, s_{\text{adv}})}{(|s_{\text{clean}}| + |s_{\text{adv}}|)/2}
\end{equation*}
where \( \Delta_\text{char}(s_{\text{clean}}, s_{\text{adv}}) \) denotes the total number of characters that differ between $s_{\text{clean}}$ and $s_{\text{adv}}$, counting insertions, deletions, and substitutions.

\noindent\textbf{Semantic Similarity.} This metric, measured by \texttt{BERTScore}, assesses how closely the adversarial text $s_{\text{adv}}$ preserves the semantics of $s_{\text{clean}}$. 

Our results in Table~\ref{tab:modification_metrics} show that character-level attacks (e.g., \textcolor{gray}{\texttt{DeepWordBug}}, \textcolor{gray}{\texttt{PuncAttack}}) introduce minimal changes, with modification rates below 1\% and BERTScore exceeding 0.97, indicating near-perfect semantic preservation. Word-level attacks such as \textcolor{gray}{\texttt{TextFooler}} and \textcolor{gray}{\texttt{BERT-Attack}} introduce slightly higher modification rates (around 7\%–14\%) while still maintaining high semantic similarity (above 0.98). In contrast, the sentence-level attack \textcolor{gray}{\texttt{StyleAdv}} results in a substantially higher modification rate (over 130\%) and a more noticeable drop in semantic similarity (BERTScore of 0.8558). This is expected, as \textcolor{gray}{\texttt{StyleAdv}} explicitly targets writing style transformation, which involves rephrasing and substituting multiple words throughout the sentence. As a result, even though the underlying meaning is largely preserved, the surface form diverges significantly from the original. These findings demonstrate that most adversarial modifications are subtle and semantically faithful—except for sentence-level style attacks, which are more aggressive in altering surface text while still maintaining reasonable semantic alignment. This confirms that the adversarial examples are both realistic and challenging to detect.

\begin{table}[ht]
\fontsize{7.1}{11}\selectfont
\centering
\caption{Quantitative comparison of text changes under different attack methods. We report both character-level \textbf{Modification Rate} and \textbf{Semantic Similarity} (BERTScore).}
\label{tab:modification_metrics}
\begin{tabularx}{0.48\textwidth}{cccc}
\toprule
Attack & Type & Modification Rate $\downarrow$ & Semantic Similarity $\uparrow$ \\
\cmidrule(lr){1-2}\cmidrule(lr){3-4}
DeepWordBug         
& Char  & 0.0090 & 0.9816 \\
PuncAttack          
& Char  & \textbf{0.0073} & 0.9749 \\
TextFooler          
& Word  & 0.0761 & \textbf{0.9885} \\
BERT-Attack         
& Word  & 0.1362 & 0.9828 \\
StyleAdv           
& Sent & 1.3117 & 0.8558 \\
\bottomrule
\end{tabularx}
\end{table}

In addition, we provide examples in Table~\ref{tab:attack_examples} to qualitatively evaluate the degree of changes. These examples show that character-level attacks (\textcolor{gray}{\texttt{DeepWordBug}}, \textcolor{gray}{\texttt{PuncAttack}}) make minimal, often imperceptible changes that nonetheless disrupt model predictions, revealing vulnerability to small spelling errors. Word-level attacks (\textcolor{gray}{\texttt{TextFooler}}, \textcolor{gray}{\texttt{BERT-Attack}}) replace words with adversarial synonyms, maintaining syntactic structure and semantics but misleading models reliant on lexical cues. Sentence-level attacks (\textcolor{gray}{\texttt{StyleAdv}}) introduce broader stylistic and phrasing changes, preserving overall meaning while posing challenges for detection.
These results highlight how different adversarial attacks vary in modification magnitude and detectability, emphasizing the need for defenses that handle perturbations from minor character edits to broader stylistic rewrites.

\begin{table*}[tb]
\footnotesize
\caption{Examples of content under different levels of textual adversarial attacks. The \textbf{Type} column indicates the attack granularity (character-, word-, or sentence-level), and the injected perturbations are highlighted in \textcolor{red}{red}.}
\small
\begin{tabularx}{\linewidth}{ccX}
\toprule
Attack & Type & \multicolumn{1}{c}{Example Sentence} \\
\cmidrule(lr){1-2}\cmidrule(lr){3-3}
Clean               & - & 	The model can also be used in conjunction with differentiable memory mechanisms for implicit relation discovery in one-shot learning tasks. \\ 
DeepWordBug         & Char & The model can also be used in conju\textcolor{red}{f}nction with differentiable memory mechanisms for implicit relation discovery in one-shot learning tasks. \\ 
PuncAttack          & Char & The model can also be used in c\textcolor{red}{\&}onjunction with differentiable memory mechanisms for implicit relation discovery in one-shot learning tasks. \\ 
TextFooler          & Word & The \textcolor{red}{paragon} can also be used in conjunction with differentiable memory mechanisms for implicit relation discovery in one-shot learning tasks. \\ 
BERT-Attack         & Word & The model can also be used in conjunction with differentiable memory mechanisms for \textcolor{red}{explicit} relation discovery in one-shot learning tasks. \\
StyleAdv            & Sent & The model can also be used in conjunction with \textcolor{red}{a variety of} differentiable memory mechanisms for the implicit relation discovery in \textcolor{red}{one shot} learning tasks. \\
\bottomrule
\end{tabularx}
\label{tab:attack_examples}
\end{table*}

\section{Components and Implementation of Textual Adversarial Attacks.} 
\label{app:attack_implementation}
Textual attacks are composed of four fundamental components: (1) \textbf{Goal Function} defines the objective of the attack, offering a heuristic for search strategies to identify optimal solutions. It evaluates how effectively perturbations achieve desired outcomes, such as convincing LLMs to favor the altered paper for better ratings and feedback. (2) \textbf{Constraints} ensure that the disturbed text maintains a resemblance to the original, preserving semantic similarity to validate the modification. (3) \textbf{Transformation} process details methods for altering input content, including character modifications, synonym replacements, and changes in style. (4) \textbf{Search Method} explores valid perturbations produced by transformations, employing techniques such as greedy search with word importance ranking, beam search, and brute-force search.

The textual attack methods we used to manipulate the paper content follow the aforementioned formulation and are primarily implemented using TextAttack\footnote{\url{https://github.com/QData/TextAttack}} \cite{morris2020textattack}. Specifically, the \textit{goal function} for all the employed attack methods-namely, \textcolor{gray}{\texttt{DeepWordBug}}, \textcolor{gray}{\texttt{PuncAttack}}, \textcolor{gray}{\texttt{TextFooler}}, \textcolor{gray}{\texttt{BERT-Attack}}, and \textcolor{gray}{\texttt{StyleAdv}} is to maximize the total score shift. For the \textit{constraint}, we basically follow the setting in PromptBench~ \cite{zhu2024promptbench} and set the max candidate (maximum number of generated perturbation candidates) of \textcolor{gray}{\texttt{TextFooler}} and \textcolor{gray}{\texttt{BERT-Attack}} to 15.

For the \textit{search} part, we adopt a greedy approach with word importance ranking for \textcolor{gray}{\texttt{DeepWordBug}}, \textcolor{gray}{\texttt{BERT-Attack}}, \textcolor{gray}{\texttt{PuncAttack}}, and \textcolor{gray}{\texttt{TextFooler}}. We restrict our process to the top 50 most important words to balance computational affordability and attack effectiveness, feeding these into our perturbation generation pipeline. 
\textcolor{gray}{\texttt{StyleAdv}} uses brute-force at the sentence level because it does not involve word-level searching. We segment paper content into individual sentences using the NLTK tokenizer\footnote{\url{https://www.nltk.org/api/nltk.tokenize.html}} \cite{bird2006nltk}. For style attacks, we utilize the pre-trained model~\cite{krishna2020reformulating} to convert text into Bible style. Original sentences are replaced with their transformed counterparts as adversarial examples.

\section{Model Configuration}
\label{app:parameters}
We accessed closed-source models (\texttt{GPT-4o} and \texttt{GPT-4o-mini}) via the OpenAI API. Open-source models were accessed through Hugging Face, including \texttt{Llama-3.3-70B} (approximately 70B parameters, running on H100 via vLLM, requiring around 140GB memory) and \texttt{Mistral-small-3.1} (approximately 24B parameters, running on H100 via vLLM, requiring around 60GB memory). All models were configured with a temperature of 0.3 to balance creativity and coherence. The maximum number of generated tokens was set to 2048, allowing detailed reviews without exceeding practical length constraints.

\section{Prompt Example}
\label{app:example_prompt}
Fig.~\ref{fig:prompt_template} presents the detailed prompt template used for automatic review generation in our experiments. The prompt is explicitly crafted to simulate the role of a professional reviewer specialized in computer science and machine learning, guiding LLMs to produce structured reviews in ICLR style. To ensure a comprehensive critique, the prompt directs the model to include both positive and negative aspects of the paper by emphasizing: ``Make sure to include both positive and negative aspects.'' This approach addresses findings from \citet{zhou2024llm}, who noted that LLMs often generate overly positive reviews, risking optimistic evaluations that overlook a paper’s weaknesses. By encouraging balanced feedback, the prompt aims to produce more critical and informative assessments.

The prompt further enhances review quality by tagging sentences with specific aspects, ensuring fine-grained analysis and differentiation of review components. It also requires the model to assign scores across multiple evaluation dimensions, with brief explanations for each, enabling quantitative assessment aligned with qualitative commentary. Additionally, to maintain conciseness and consistency, we specified that the total output should not exceed 500 tokens. This design promotes consistent, reproducible review generation that mirrors real-world peer review processes, allowing us to systematically evaluate the robustness and interpretability of LLM-based reviewing systems.

\begin{figure*}[t]
\centering
\begin{tcolorbox}[
  colback=gray!5,
  colframe=gray!180,
  title=Prompt Template,
  fonttitle=\bfseries,
  boxrule=0.5pt,
  arc=2mm,
  left=2mm, right=2mm, top=1mm, bottom=1mm
]
You are a professional reviewer in computer science and machine learning. Based on the given content of a research paper, you need to write a review in ICLR style and tag sentences with the corresponding tag type at the beginning of sequence: tags types: [NONE], [SUMMARY], [MOTIVATION POSITIVE], [MOTIVATION NEGATIVE], [SUBSTANCE POSITIVE], [SUBSTANCE NEGATIVE], [ORIGINALITY POSITIVE], [ORIGINALITY NEGATIVE], [SOUNDNESS POSITIVE], [SOUNDNESS NEGATIVE], [CLARITY POSITIVE], [CLARITY NEGATIVE], [REPLICABILITY POSITIVE], [REPLICABILITY NEGATIVE], [MEANINGFUL COMPARISON POSITIVE], [MEANINGFUL COMPARISON NEGATIVE]. Your total output should not surpass 500 tokens, make sure to include both positive and negative aspects. Also, you need to predict the review score in several aspects based on the generated review, providing an explanation of each aspect in less than 30 tokens. Choose an integer score from 1 to 10, higher score means better paper quality. 
\\

Section \{section name 1\}: \{section content 1\} \\
Section \{section name 2\}: \{section content 2\} \\
... \\
Section \{section name N\}: \{section content N\}
\\

Please strictly follow the format of Example output: 1. REVIEW: tagged sequences. 2. REVIEW SCORE: OVERALL: score, SUBSTANCE: score, APPROPRIATENESS: score, MEANINGFUL COMPARISON: score, SOUNDNESS CORRECTNESS: score, ORIGINALITY: score, CLARITY: score, IMPACT: score. 3. REVIEW SCORE EXPLANATION: OVERALL: explanation, SUBSTANCE: explanation, APPROPRIATENESS: explanation, MEANINGFUL COMPARISON: explanation, SOUNDNESS CORRECTNESS: explanation, ORIGINALITY: explanation, CLARITY: explanation, IMPACT: explanation.
\end{tcolorbox}
\caption{Prompt template used for review generation.}
\label{fig:prompt_template}
\end{figure*}

\section{Experiment Result of PeerRead Dataset}
\label{app:peerread_attack_result}
\subsection{Evaluation -- Review Quality}
The evaluation of review quality of PeerRead dataset is provided in Table~\ref{tab:quality_result_peerread}, which exhibits trends consistent with those discussed in Section~\ref{sec:evaluation_review_quality}.

\begin{table}[ht]
\fontsize{7.1}{11}\selectfont
\caption{Evaluation of LLM-generated reviews' Comprehensiveness and Similarity to humans of PeerRead~\cite{kang-etal-2018-dataset} dataset. ACOV denotes Aspect Coverage. R-1, R-2, R-L, and BS indicate the Rouge-1/2/L, and BertScore, respectively. \textbf{Human (within)} reports the mean scores of within-paper pairs, \textbf{Human (across)} reports the mean over across-paper pairs.}
\label{tab:quality_result_peerread}
\begin{tabularx}{0.5\textwidth}{Xccccc}
\toprule
Reviewer & ACOV & R-1 & R-2 & R-L & BS \\
\cmidrule(lr){1-1}\cmidrule(r){2-6}
Human (within) & \textbf{0.5285} & \textbf{0.4148} & \textbf{0.0973} & \textbf{0.1899} & \textbf{0.8403} \\
Human (across) & \textbf{0.5285} & 0.2188 & 0.0285 & 0.1154 & 0.8179 \\
\cmidrule(lr){1-1}\cmidrule(r){2-6}
GPT-4o-mini & 0.6066 & \textbf{0.3938} & 
0.0676 & \textbf{0.1875} & \textbf{0.8341} \\
GPT-4o & 0.9151 & 0.3469 & 0.0678 & 0.1676 & 0.8251 \\
Llama-3.3-70B & 0.6489 & 0.3432 & \textbf{0.0757} & 0.1760 & 0.8208 \\
Mistral-small-3.1 & \textbf{0.9202} & 0.3541 & 0.0737 & 0.1658 & 0.8220  \\
\bottomrule
\end{tabularx}
\end{table}

\subsection{Evaluation -- Attack Effectiveness}
\begin{table*}[tb]
\caption{Adversarial Attack Performance and Score Shift Comparison of PeerRead~\cite{kang-etal-2018-dataset} dataset. ``ASR'' represents the Attack Success Rate. ``Avg. Score shift'' signifies the average score shift between attacked outputs and clean outputs. ``Avg. \#Pos/Neg Shift'' indicates the average change in the number of positive and negative tags assigned to the reviews. ``\#Queries'' refers to the average number of times LLM is queried. The values are reported as result from \texttt{GPT-4o-mini} / \texttt{GPT-4o}/ \texttt{Llama-3.3-70B} / \texttt{Mistral-small-3.1}.}
\label{tab:adv_result_peerread}
\small
\begin{center}
\resizebox{\textwidth}{!}{
\setlength{\tabcolsep}{5pt}
\begin{tabular}{cccccc}
\toprule
\multirow{2}{*}[-3pt]{Attacks} &
\multirow{2}{*}[-3pt]{ASR $\uparrow$} &
\multicolumn{3}{c}{Average Shift} &
\multirow{2}{*}[-3pt]{\#Queries $\downarrow$} \\
\cmidrule(lr){3-5}
 &  &  
 Score $\uparrow$ & 
 \#Pos Tags $\uparrow$ & 
 \#Neg Tags $\downarrow$ & 
 \\
\cmidrule(lr){1-1}\cmidrule(r){2-6}
DeepWordBug 
& \textbf{0.92} / 0.81 / 0.82 / 0.89  
& 4.27 / 3.87 / 3.06 / 3.22  
& 0.33 / 0.00 / \textbf{0.07} / -0.02  
& \textbf{-0.95} / -1.18 / -1.73 / -0.24  
& 42 / 51 / 50 / 56 
\\
PuncAttack 
& 0.87 / 0.90 / 0.79 / 0.83   
& 3.90 / 4.23 / \textbf{3.17} / 2.98 
& 0.33 / -0.03 / 0.03 / -0.02  
& -0.81 / -1.18 / -1.71 / -0.20  
& 51 / 63 / 64 / 68  
\\
TextFooler 
& \textbf{0.92} / \textbf{0.95} / \textbf{0.87} / \textbf{0.91}  
& \textbf{4.61} / 4.30 / 2.87 / 3.44  
& \textbf{0.35} / -0.03 / -0.07 / \textbf{-0.01}   
& -0.90 / \textbf{-1.50} / \textbf{-2.14} / -0.32  
& 46 / 54 / 57 / 58  
\\
BERT-Attack 
& 0.85 / 0.74 / 0.70 / 0.85
& 4.13 / \textbf{4.43} / 2.35 / \textbf{3.64}  
& 0.28 / \textbf{0.03} / -0.09 / -0.02  
& -0.75 / -1.23 / -1.70 / \textbf{-0.40}  
& 52 / 53 / 54 / 65  
\\
StyleAdv 
& 0.83 / 0.88 / 0.80 / 0.87  
& 3.58 / 4.28 / 2.35 / 2.81  
& 0.27 / -0.10 / -0.14 / -0.02  
& -0.65 / -1.13 / -1.96 / -0.26  
& \textbf{12} / \textbf{16} / \textbf{12} / \textbf{16} 
\\
\bottomrule
\end{tabular}
}
\end{center}
\end{table*}
Table~\ref{tab:adv_result_peerread} shows the results of different attacks designed to boost scores assigned by the LLM-based reviewer. The trends align with those in Table~\ref{tab:adv_result}, demonstrating the attacks’ effectiveness across datasets and the vulnerability of LLMs to simple textual adversarial attacks in a black-box setting. A closer analysis reveals that: \\
\noindent\textbf{TextFooler effectively manipulates LLM evaluations across models.} It consistently achieves the highest ASR, peaking at 0.95 for \texttt{GPT-4o}, showing that even minor lexical substitutions can significantly distort judgments. In general, positive tags tend to increase relative to negative tags, with the decrease in negative tags usually larger than the increase in positive tags. This biases reviews toward favorable evaluations and suppresses critical ones, undermining review integrity. The pattern does not hold for every model, as prompts that encourage both positive and negative feedback and emphasize critical evaluation can cause models that already assign high positive and low negative scores to reduce positive tags or increase negative tags after the attack, reflecting the combined effect of the prompt and the perturbation.\\
\noindent\textbf{StyleAdv is the most efficient attack among the tested methods.} Despite achieving only moderate ASR, it produces noticeable score shifts with the lowest query count. This indicates that stylistic manipulations, even with minimal transformation candidates and limited model interaction, can effectively deceive LLMs. Their high modification rate allows the attacked content to differ sufficiently from the original, altering its distribution in a way that impacts evaluations.\\
\noindent\textbf{GPT models are susceptible to adversarial attacks, while Llama-3.3-70B shows better resilience.} We hypothesize that this difference stems from GPT models’ stronger emphasis on maintaining contextual coherence. This strength in natural text generation makes GPT models more likely to interpret even manipulated sentiment shifts as coherent opinions, resulting in more pronounced score changes. In contrast, \texttt{Llama-3.3-70B} may prioritize semantic understanding over coherence, which provides slightly greater robustness against such manipulations.

Overall, these results underscore the importance of advancing LLM robustness and point to resilient prompting and processing strategies as promising directions for safeguarding review integrity against adversarial perturbations.

\section{Human Evaluation}
We conduct a controlled human evaluation to complement automatic metrics and assess the quality and robustness of LLM-generated peer reviews. 
This evaluation is structured around two key research questions:\\
\textit{(1) How similar are LLM-generated reviews to human-written ones in terms of perceived quality?}\\
\textit{(2) To what extent adversarial textual patterns of varying granularity can remain stealthy when embedded in paper content.}\\
To answer these questions, we designed two complementary human studies: one focused on review quality evaluation, and another on attack stealthiness and identifiability.
We collected user study results from 29 participants, all of whom are master's or PhD students with background knowledge in deep learning and natural language processing.
The setup and methodology for each in turn are described below.

\subsection{Review Quality Ranking}
\label{app:human_evaluation_review_quality}
While automatic evaluation metrics offer a broad assessment of LLM-generated reviews, demonstrating their ability to cover aspects and exhibit high semantic similarity to human-written reviews, they fall short in capturing the qualities of review usefulness, coherence, and perceived value from a human perspective. Specifically, such metrics may favor reviews that are grammatically fluent and lexically aligned with human-written text, even when such reviews lack substantive critique or meaningful engagement with the paper. To mitigate these limitations, we conduct a controlled human evaluation to complement automatic analysis and offer a more comprehensive understanding of review quality. 
\subsubsection{Evaluation Criteria}
To assess review quality as perceived by humans, we employ the following three criteria: \\
(1) \textbf{Fluency \& Consistency:} \textit{Is the review easy to read and internally coherent?} \\
(2) \textbf{Content Relevance:} \textit{Does the review address the paper's core contributions and align with the ground-truth meta-review context?} \\
(3) \textbf{Insightfulness:} \textit{Does the review provide substantive, constructive feedback that could help improve the paper?} 
\subsubsection{Annotation Process.}
To keep the task manageable and avoid annotator fatigue, we presented the meta-review of each paper, which served as a concise conference-style summary instead of the full manuscript. This meta-review provided the context for evaluating the relevance, coherence, and insightfulness of the reviews. 

Participants assessed bundles of four reviews, each consisting of one LLM-generated review (from \texttt{GPT-4o-mini}, \texttt{GPT-4o}, \texttt{Llama-3.3-70B}, or \texttt{Mistral-small-3.1}) and three human-written reviews. Every bundle was independently judged by all 29 participants. For each bundle, annotators ranked the four anonymized reviews along the three quality dimensions. After completing each ranking task, they also reported their confidence on a 1–5 scale, reflecting the certainty of their judgments. 
\subsubsection{Result \& Discussion.}
The human evaluation results in Table~\ref{tab:user_study_part1} reveal insights emerging regarding the human-perceived quality of LLM-generated reviews as follows:\\
\noindent\textbf{\texttt{Llama-3.3-70B} consistently excels in fluency and content relevance.} It secured the highest rankings (lowest \textit{Rank} values) in both \textit{Fluency \& Consistency} and \textit{Content Relevance}. While its \textit{Insightfulness} was slightly below that of \texttt{GPT-4o-mini}, it remained competitive and outperformed \texttt{GPT-4o} across most dimensions, highlighting its overall strength as a balanced review generator.

\noindent\textbf{\texttt{GPT-4o} underperforms in insightfulness and overall constructive feedback.} It showed weaker performance across all three criteria compared to other LLMs, with the lowest T-1 Rate and highest average Rank indicating difficulty in providing in-depth, constructive feedback. This limitation is not simply due to model size, as both \texttt{GPT-4o-mini} and \texttt{Llama-3.3-70B} performed better. The likely explanation is \texttt{GPT-4o}'s strict adherence to the structured prompt: while this ensured balanced coverage, it often produced exhaustive but surface-level comments, reducing sharp, focused critique. In contrast, \texttt{GPT-4o-mini} followed the prompt less rigidly, generating more targeted and insightful feedback, which human judges rewarded with higher T-1 Rates. Despite its shortcomings, \texttt{GPT-4o} still outperformed at least one human-written review in ranking tasks, demonstrating overall competitiveness.

\noindent\textbf{\texttt{Mistral-small-3.1} excels in fluency but lacks insightfulness.} Although it achieved the highest T-1 Rate for \textit{Fluency \& Consistency}, it lagged in \textit{Insightfulness}, suggesting that fluency alone does not guarantee meaningful critique.

\noindent\textbf{\texttt{GPT-4o-mini} achieves balanced and insightful performance across criteria.} It delivered competitive results in all evaluated dimensions, particularly in \textit{Insightfulness}, indicating that smaller, more efficient models can still achieve strong human-perceived review quality.

These results collectively highlight the importance of balancing fluency, relevance, and depth in LLM-generated reviews, with \texttt{Llama-3.3-70B} setting a strong benchmark in this regard. However, the somewhat underwhelming performance of models like \texttt{GPT-4o} in providing insightful feedback underscores the ongoing challenge of enhancing LLMs to generate not just coherent text but also substantive and contextually valuable content.

\begin{table*}[tb]
\caption{User Study on Review Quality Ranking. ``Rank'' denotes the average position of the LLM-generated review among the three human reviews (lower is better), and ``T-1 Rate'' is the proportion of times the LLM review was ranked first for that criterion. ``Conf.'' is the average confidence score users assigned to their rankings.}
\label{tab:user_study_part1}
\footnotesize
\begin{center}\
\resizebox{\textwidth}{!}{%
\begin{tabular}{ccccccccccc}
\toprule
\multirow{2}{*}[-3pt]{Attacks} &
\multicolumn{3}{c}{Fluency \& Consistency} &
\multicolumn{3}{c}{Content Relevance} &
\multicolumn{3}{c}{Insightfulness}
\\
\cmidrule(lr){2-4}\cmidrule(lr){5-7}\cmidrule(lr){8-10}
&
Rank $\downarrow$ & T-1 Rate $\uparrow$ & Conf. $\uparrow$ &
Rank $\downarrow$ & T-1 Rate $\uparrow$ & Conf. $\uparrow$ &
Rank $\downarrow$ & T-1 Rate $\uparrow$ & Conf. $\uparrow$ &
 \\
\cmidrule(lr){1-1}\cmidrule(r){2-11}
GPT-4o-mini
& 2.07 & 0.45 & \textbf{4.03} & 1.83 & 0.52 & \textbf{4.03} & \textbf{1.66} & \textbf{0.55} & \textbf{4.03} \\
GPT-4o
& 2.31 & 0.31 & 3.93 & 2.31 & 0.38 & 3.93 & 2.66 & 0.17 & 3.93 \\
Llama-3.3-70B
& \textbf{1.90} & 0.45 & 3.97 & \textbf{1.69} & \textbf{0.59} & 3.97 & 1.72 & 0.52 & 3.97 \\
Mistral-small-3.1 
& 2.10 & \textbf{0.52} & 3.72 & 1.93 & 0.41 & 3.72 & 2.41 & 0.24 & 3.72 \\
\bottomrule
\end{tabular}%
}
\end{center}
\end{table*}

\subsection{Attack Imperceptibility Evaluation}
\label{app:human_evaluation_attack_imperceptibility}
To further examine the robustness of LLM-generated content, we conducted a human evaluation targeting the identifiability of adversarial perturbations under realistic review scenarios. 
\subsubsection{Evaluation Criteria.} 
Specifically, this experiment aimed to answer two key research questions: \\
(1) \textbf{Attack Identification:} \textit{In a simulated peer-review setting, can human evaluators distinguish which papers have been adversarially attacked?} \\
(2) \textbf{Attack Localization:} \textit{When explicitly informed that the paper content has undergone an adversarial attack, can evaluators locate the specific paragraph containing the adversarial perturbation?}
\subsubsection{Annotation Process.} 
We recruited annotators with prior experience in academic writing to participate in a two-stage evaluation. In the first stage, annotators were presented with pairs of papers, one original and one adversarially perturbed, and were asked to determine which paper had been manipulated. In the second stage, annotators were provided with a single adversarially perturbed paper and were instructed to identify the paragraph they judged to be most indicative of adversarial modification under a specified textual attack method.

\setlength{\tabcolsep}{4.5pt} 
\begin{table}[!t]
\fontsize{7.1}{11}\selectfont
\caption{User Study of Attack Imperceptibility. The study consists of two parts: \textit{Attack Identification} and \textit{Attack Localization}. ``Acc.'' represents the accuracy of identifying the attacked paper or position, ``Unc.'' denotes the \textit{Uncertainty Rate}, calculated as the proportion of times users responded with ``N/A, can’t distinguish.'' ``Conf.'' refers to the average confidence score users assigned to their answers.}
\label{tab:user_study_part2}
\begin{tabularx}{0.5\textwidth}{ccccccc}
\toprule
\multirow{2}{*}[-3pt]{Attacks} &
\multicolumn{3}{c}{Identification} & 
\multicolumn{3}{c}{Localization} 
\\
\cmidrule(lr){2-4}\cmidrule(r){5-7}
&
Acc. $\downarrow$ & Unc. $\uparrow$ & Conf. $\downarrow$ &
Acc. $\downarrow$ & Unc. $\uparrow$ & Conf. $\downarrow$
\\
\cmidrule(lr){1-1}\cmidrule(r){2-7}
DeepWordBug 
& 0.62 & 0.55 & \textbf{3.28} & 0.96 & 0.14 & 4.31 
\\
PuncAttack  
& 0.88 & 0.41 & 4.10 & 0.69 & 0.10 & 4.48
\\
TextFooler  
& 0.63 & 0.45 & 3.45 & 0.26 & 0.07 & 3.17
\\
BERT-Attack 
& \textbf{0.50} & \textbf{0.72} & 3.48 & \textbf{0.08} & 0.10 & 2.86
\\
StyleAdv    
& 0.73 & 0.24 & 3.76 & 0.21 & \textbf{0.34} & \textbf{2.79}
\\
\bottomrule
\end{tabularx}
\end{table}

This setting reflects real-world reviewing conditions where reviewers rely solely on textual cues. The evaluation helps assess not only the perceptibility of adversarial patterns to informed readers but also whether these attacked papers can effectively evade human detection even under heightened scrutiny, highlighting the potential risk of such subtle attacks in influencing LLM-based review systems.

\subsubsection{Result \& Discussion.}
As shown in Table~\ref{tab:user_study_part2}, the imperceptibility of adversarial attacks varies substantially across different granularities, as discussed below. 

\noindent\textbf{\texttt{BERT-Attack} is the most stealthy adversarial method.} Annotators achieved only 50\% accuracy in identifying perturbed papers (i.e., chance level) and a low localization accuracy of 8\%. This indicates that semantic-preserving synonym substitution strategies are highly effective in maintaining naturalness under human scrutiny, making adversarial changes difficult to detect even when readers are explicitly aware of their presence.

\noindent\textbf{\texttt{StyleAdv} is easier to identify but difficult to localize.} Although it was designed to subtly alter writing style, it showed relatively high identification accuracy (73\%) yet poor localization accuracy (21\%), along with the highest uncertainty rate (34\%). This suggests that stylometric perturbations do not blend well in academic contexts, likely because injected styles (e.g., biblical) deviate too far from scholarly tone. As detailed in Appendix~\ref{app:attackExample_modificationRate}, \textcolor{gray}{\texttt{StyleAdv}} also involves higher modification rates and lower semantic similarity, which may disrupt paragraph-level cohesion and topical consistency, making manipulations stand out when compared side-by-side with a clean version. However, the poor localization accuracy indicates that in zero-shot settings without a clean reference, humans struggle to pinpoint the attack, highlighting its stealthiness. This risk could grow if future attacks adopt writing styles more plausible in academic contexts, such as those resembling non-native writing patterns.

\noindent\textbf{\texttt{PuncAttack} leaves obvious patterns yet hinders accurate localization.} It yielded high identification accuracy, which is expected since even minimal punctuation edits become salient when compared with a clean version. However, its moderate localization accuracy suggests that while readers sense something unusual, they struggle to pinpoint the exact manipulated paragraph, especially when subtle noise is sparsely distributed across a lengthy article. In real-world settings without prior knowledge of an attack, such perturbations may be dismissed as typos or stylistic quirks, allowing them to evade scrutiny. Despite its simplicity, \textcolor{gray}{\texttt{PuncAttack}} retains strong potential as a subtle and deceptive adversarial strategy.

\noindent\textbf{\texttt{TextFooler} and \texttt{DeepWordBug} occupy the middle ground but differ in detectability.} \textcolor{gray}{\texttt{DeepWordBug}} avoids obvious cues such as punctuation, making its character-level perturbations hard to detect across lengthy paper content, which results in relatively low identification accuracy. Yet when annotators know attacks are present and examine carefully, localization accuracy rises to 96\%, showing that spelling changes can be discovered with focused reading. In contrast, \textcolor{gray}{\texttt{TextFooler}} exhibits lower identification and localization accuracies, suggesting that synonym substitution attacks are harder to detect due to higher semantic preservation, which reduces perceptibility.

These findings show that attack detectability depends heavily on nature and granularity: semantic-preserving attacks are more stealthy, while stylistic or punctuation anomalies are easier to spot in side-by-side comparisons yet remain risky in real-world, zero-knowledge settings. This highlights the challenge of building robust defenses against diverse, evolving attack methods.

\section{Ablation Study -- Impact of Aspect Tags in Prompts on ACOV Scores}
\label{app:abs_acov}

Considering the ACOV advantage may stem from the structured prompt, in particular, explicit aspect tags may guide models to address each dimension systematically, thereby increasing coverage relative to human-written reviews. To assess the extent of this effect, we perform an ablation study in which aspect tags are either included or removed from the prompts. Results are reported in Table~\ref{tab:abs_acov}.

\begin{table}[h]
\caption{Ablation study on the PeerRead dataset examining the impact of explicit aspect tags in prompts on ACOV scores. ``w/ tag'' and ``w/o tag'' denote prompts with and without explicit aspect tags, respectively. ``Llama'' refers to Llama-3.3-70B, and ``Mistral'' to Mistral-small-3.1.}
\label{tab:abs_acov}
\small
\begin{center}
\resizebox{0.48\textwidth}{!}{%
\setlength{\tabcolsep}{6pt}
\begin{tabularx}{0.48\textwidth}{ccccc}
\toprule
\textbf{Models} & \textbf{GPT-4o-mini} & \textbf{GPT-4o} & \textbf{Llama} & \textbf{Mistral} \\
\cmidrule(lr){1-1}\cmidrule(r){2-5}
w/ tag & 
\textbf{0.6066} &
\textbf{0.9154} & \textbf{0.6489} &
\textbf{0.9202} \\
w/o tag & 
0.3556 &
0.4317 & 0.1702 &
0.4024 \\
\bottomrule
\end{tabularx}%
}
\end{center}
\end{table}

Across all models, the removal of explicit aspect tags leads to a decrease in ACOV. This indicates that structured prompt design plays a central role in achieving higher coverage. Consistent with prior work~\cite{yuan2022can}, we emphasize that ACOV should be interpreted as a coverage-oriented metric, not as a direct indicator of human-level review quality. The ablation highlights that, when explicitly guided, LLMs are able to systematically cover the intended review template, underscoring their potential as tools for generating structured and comprehensive feedback.

\section{Ablation Study -- Impact of Attack Focus Localization}
To evaluate the contribution of the proposed Attack Focus Localization algorithm, we conduct an ablation study that examines its role in improving adversarial effectiveness on long scientific documents. Unlike previous black-box adversarial methods that primarily target short inputs, the proposed approach identifies critical regions within lengthy academic papers and applies perturbations selectively. This localized strategy allows perturbations to be concentrated on semantically important sections, thereby enhancing attack effectiveness.

Table~\ref{tab:afl_ablation} presents the results of this ablation study on the PeerRead dataset. For each attack method, we compare the \textit{Attack Success Rate (ASR)} and the \textit{Average Score Shift} when perturbations are applied to the entire document without localization versus when using our proposed strategy.

\begin{table}[h]
\caption{Ablation study on the impact of Attack Focus Localization on adversarial effectiveness. Values are presented as (Without/With Attack Focus Localization).}
\label{tab:afl_ablation}
\small
\begin{center}\
\resizebox{0.48\textwidth}{!}{%
\setlength{\tabcolsep}{10pt}
\begin{tabularx}{0.5\textwidth}{ccc}
\toprule
\textbf{Attacks} & \textbf{ASR $\uparrow$} & \textbf{Average Score Shift $\uparrow$} \\
\cmidrule(lr){1-1}\cmidrule(r){2-3}
DeepWordBug & \textbf{0.11} / \textbf{0.92} & -0.12 / 4.27 \\
PuncAttack  & 0.10 / 0.87 & -0.11 / 3.90 \\
TextFooler  & 0.10 / \textbf{0.92} &  \textbf{0.22} / \textbf{4.61} \\
BERT-Attack & 0.10 / 0.85 &  0.02 / 4.13 \\
StyleAdv    & 0.10 / 0.83 & -0.10 / 3.58 \\
\bottomrule
\end{tabularx}%
}
\end{center}
\end{table}

The results show that incorporating the proposed Attack Focus Localization algorithm consistently boosts both ASR and Average Score Shift across all attack methods. These findings demonstrate that this localization strategy serves as a generalizable and effective approach for adversarial attacks on long-form academic documents, highlighting its importance in evaluating the robustness of LLM-based peer review systems.

\end{document}